# Reliability-Aware CT–MRI Registration: A Quality Engineering Framework with Stability Analysis and Risk Classification

**Nisreen Albzour**

*School of Systems Science and Industrial Engineering, Binghamton University, Binghamton, NY, USA*

ORCID: https://orcid.org/0009-0000-9317-340X

## Abstract

Multimodal CT–MRI registration underpins image-guided radiotherapy, surgical navigation, and diagnostic workflows, yet existing pipelines report only aggregate quality metrics and provide no per-case reliability signal to support clinical decisions. A registration that converges to a plausible but incorrect local optimum can still score well on conventional post-hoc metrics, creating a silent failure mode. We propose a reliability-aware framework that converts continuous per-case registration quality into actionable Green/Yellow/Red risk categories using data-learned thresholds, enabling automated acceptance, expert-review triggering, and rejection within one pipeline. CT was registered to T1-weighted MRI using rigid and affine transformations on 90 paired slices from 18 patients (14 training, 4 held-out test) across brain (n=9), abdominal (n=8), and neck (n=1) anatomies. Reliability was assessed through quality improvement (ΔNMI, ΔSSIM), spatial overlap (Dice), registration stability across repeated initialisations, and inverse consistency error, combined into a single score R. Thresholds learned from training patients were applied unchanged to test patients. Affine registration significantly outperformed rigid on NMI (Cohen's $d = 0.76$, $p < 10^{-13}$) and SSIM ($d = 0.36$, $p < 0.01$), yielding 44% Green (automatically acceptable) classifications versus 33% for rigid. Reliability-filtered registrations surpassed unfiltered methods on all metrics. Per-anatomy performance varied markedly (abdominal 73% Green, brain 22%), and weight-sensitivity analysis identified Dice overlap as the dominant reliability component. This interpretable, data-driven framework provides a reproducible foundation for quality-controlled multimodal registration and applies to any rigid or affine algorithm. Risk thresholds reflect statistical rather than clinical validation, which should precede clinical deployment.

***Keywords:*** *Image registration, CT–MRI, quality assurance, reliability engineering, risk stratification, inverse consistency, registration stability*

## 1. Introduction

Computed tomography (CT) and magnetic resonance imaging (MRI) provide complementary anatomical information: CT offers high spatial resolution and bone contrast, while MRI excels at soft-tissue differentiation. Combining these modalities through image registration is foundational to radiotherapy planning, surgical navigation, and multimodal diagnostic workflows [1]. However, registration is itself an estimation problem with no ground truth available at deployment time, and a misregistered image pair can silently propagate errors into downstream clinical decisions [2].

Despite decades of methodological development in image registration—from intensity-based similarity metrics [3] to deep-learning-based deformable registration [4]—the great majority of registration pipelines report only aggregate quality metrics (e.g., mean mutual information or Dice score across a dataset) without providing a per-case indication of whether a specific registration result can be trusted [5]. This is a critical gap from a quality-engineering perspective: a registration pipeline deployed in practice does not have access to ground-truth alignment, and therefore needs an internal signal for when a result is reliable enough to accept automatically, when it warrants expert review, and when it should be rejected and re-attempted [6].

Reliability engineering principles—which emphasise quantifying confidence in a system's output rather than reporting only point estimates—have been increasingly recognised as essential for trustworthy deployment of imaging algorithms [7]. Yet registration reliability differs fundamentally from the uncertainty quantification problem in reconstruction or correction tasks: there is no probabilistic forward model linking a transformation parameter to image intensity, and the relevant failure modes (local minima in the optimisation landscape, inconsistent convergence across initialisations, and asymmetric forward/backward registration) require different diagnostic tools than posterior sampling or confidence intervals.

In this work, we address this gap by proposing a reliability-aware CT–MRI registration framework grounded in four complementary diagnostics: (i) quality improvement, measured via the change in normalised mutual information and structural similarity before and after registration; (ii) spatial overlap, measured via the Dice coefficient between body masks; (iii) registration stability, measured by repeating registration from multiple random initialisations and quantifying the variation in the resulting transformation parameters; and (iv) inverse consistency, measured by comparing forward (CT→MRI) and backward (MRI→CT) registration and computing the residual discrepancy after a round trip. These four diagnostics are combined into a single reliability score R, which is mapped to a three-tier risk classification (Green, Yellow, Red) using thresholds learned exclusively from training patients and validated on held-out test patients. This study uses the paired CT–MRI dataset as an independent experimental setting and addresses registration reliability rather than attenuation correction.

**Our contributions are as follows:**

**(1) A reliability score for image registration** that combines four independently meaningful diagnostics (quality improvement, overlap, stability, inverse consistency) into a single, interpretable, weighted score, rather than relying on any single post-registration metric in isolation.

**(2) A traffic-light risk classification system** with thresholds learned from training data and validated on held-out test patients, providing an actionable decision rule (accept / review / repeat) suitable for quality-controlled imaging pipelines.

**(3) A registration stability and inverse consistency diagnostic** that quantifies registration trustworthiness without requiring ground-truth alignment, using only repeated optimisation and forward–backward consistency—diagnostics that are computable at deployment time on any new patient.

**(4) A systematic, statistically validated evaluation** across 18 patients spanning three anatomies (brain, abdominal, neck), including baseline comparison against unfiltered rigid and affine registration, per-anatomy analysis, and a weight sensitivity study identifying which reliability components drive the score.

The remainder of this paper is organised as follows. Section 2 reviews related work in image registration, registration quality assessment, and reliability engineering for medical imaging. Section 3 presents the methodology. Section 4 reports results across all stages of the pipeline. Section 5 discusses findings and limitations. Section 6 concludes.

## 2. Related Work

This section situates the proposed framework relative to three bodies of literature. Section 2.1 reviews classical and learning-based approaches to multimodal CT–MRI registration. Section 2.2 examines how registration quality is conventionally assessed, including the role of inverse consistency as an internal validity check. Section 2.3 reviews reliability engineering and uncertainty quantification principles in medical imaging more broadly, including risk-stratified, traffic-light decision frameworks. Section 2.4 summarises the resulting gap that motivates this work.

### 2.1 Multimodal Medical Image Registration

Intensity-based registration methods align images by optimising a similarity metric over a space of geometric transformations. Mattes mutual information [3] remains a standard choice for multimodal registration because it does not assume a linear intensity relationship between modalities, making it well suited to CT–MRI alignment where tissue contrast differs substantially between scans. Rigid registration (translation and rotation) is the most constrained and most stable transformation class, while affine registration (adding scaling and shearing) offers greater flexibility at the cost of additional parameters to estimate [8]. More recently, deformable and deep-learning-based registration methods have been proposed to capture non-linear anatomical variation [4], though these typically require larger training datasets and provide less interpretable failure modes than classical rigid or affine registration; a comprehensive survey of this transition, including the growing role of uncertainty estimation in learning-based registration, is provided by Chen et al. [9].

Our work uses classical rigid and affine registration via Mattes mutual information, consistent with established practice for CT–MRI alignment in radiotherapy and neuroimaging [1], but shifts the contribution from the registration algorithm itself to the reliability assessment layered on top of it.

## 2.2 Registration Quality Assessment

Registration quality is conventionally assessed using similarity metrics such as mutual information, normalised cross-correlation, or structural similarity computed between the registered and reference images [10]. Dice overlap between segmented structures provides a complementary, anatomically grounded measure of alignment when masks or segmentations are available [11]. However, these metrics share a common limitation: they describe how well two specific images agree post-registration, but they do not characterise how reliable the registration process itself was, nor whether the same algorithm would produce a consistent result under a different initialisation. A registration that achieves high post-hoc similarity by converging to a plausible but incorrect local optimum would still score well on these metrics.

Inverse consistency—the requirement that registering image A to B and then B to A should approximately return the original image—has been used as an internal consistency check for deformable registration algorithms [12], with subsequent work formalising inverse consistency error as an explicit, mutual-information-driven optimisation and evaluation criterion [13], but is less commonly incorporated into rigid and affine registration pipelines despite being computable without any additional data. Our framework explicitly incorporates inverse consistency as one of four reliability components, alongside a stability diagnostic that, to our knowledge, has not been systematically combined with inverse consistency and quality metrics into a single actionable score for CT–MRI registration.

## 2.3 Reliability, Uncertainty, and Quality Engineering in Medical Imaging

Reliability engineering frameworks emphasise quantifying confidence in system outputs and using that confidence to drive decision rules—accepting, flagging, or rejecting a result—rather than treating every output as equally trustworthy [7]. In medical imaging, this principle has been applied primarily to detection and segmentation tasks, where calibrated uncertainty estimates can flag regions of low confidence for clinician review [14]. Quality improvement methodologies in healthcare imaging similarly advocate for explicit, auditable decision criteria rather than relying on practitioner judgement alone [15], and decision-making under uncertainty is recognised as a pervasive feature of clinical workflows that benefits from structured, quantitative support [16]. This emphasis is consistent with prior reliability-aware medical-AI studies that prioritise train/test discipline, external validation, calibration, and decision-relevant outputs over raw accuracy, including systematic optimisation and statistical validation of vision transformers for cervical cancer classification [17,18,19], a leakage-aware comparative benchmark of machine learning, deep learning, and transformer models for leukemia detection [20,21], studies of cross-cohort external validation, calibration, and explainability transfer for breast cancer survival prediction [22,23], and feature-selection-driven machine learning for clinical outcome prediction in post-stroke rehabilitation [24].

Risk-stratified classification systems—commonly implemented as traffic-light schemes—are widely used in process control and quality engineering to translate continuous risk scores into actionable categories with different response protocols [25]. Learning decision thresholds directly from data, rather than fixing them arbitrarily, has similarly been advocated in clinical risk-stratification settings as a way to ground operating points in observed practice rather than ad hoc convention [26], an approach our train/test threshold-learning procedure mirrors for registration reliability. Our framework adapts this paradigm to registration quality control: rather than presenting a continuous reliability score alone, we translate it into Green/Yellow/Red categories with thresholds learned from training data, enabling automated acceptance, expert review triggering, and rejection within a single, reproducible decision pipeline.

### 2.4 Summary of Literature Gap

The literature on CT–MRI registration is mature in algorithmic terms but underdeveloped in reliability assessment: existing pipelines report aggregate post-registration quality but rarely provide a per-case, decision-actionable reliability signal that combines quality, overlap, stability, and consistency. Our framework addresses this gap directly, providing a quality-engineering layer that can be applied on top of any rigid or affine registration algorithm, validated with the same train/test discipline and statistical rigour expected in quantitative engineering research.

## 3. Methodology

This study performs slice-level 2D CT–MRI registration on paired axial CT and T1-MRI images rather than full volumetric 3D registration. The proposed framework proceeds through three stages: (Stage 0) dataset verification and slice selection; (Stage 1) registration, quality evaluation, stability testing, inverse consistency, and reliability scoring; and (Stage 2) baseline comparison, statistical validation, per-anatomy analysis, weight sensitivity, and threshold validation. Figure 1 summarises the complete pipeline.

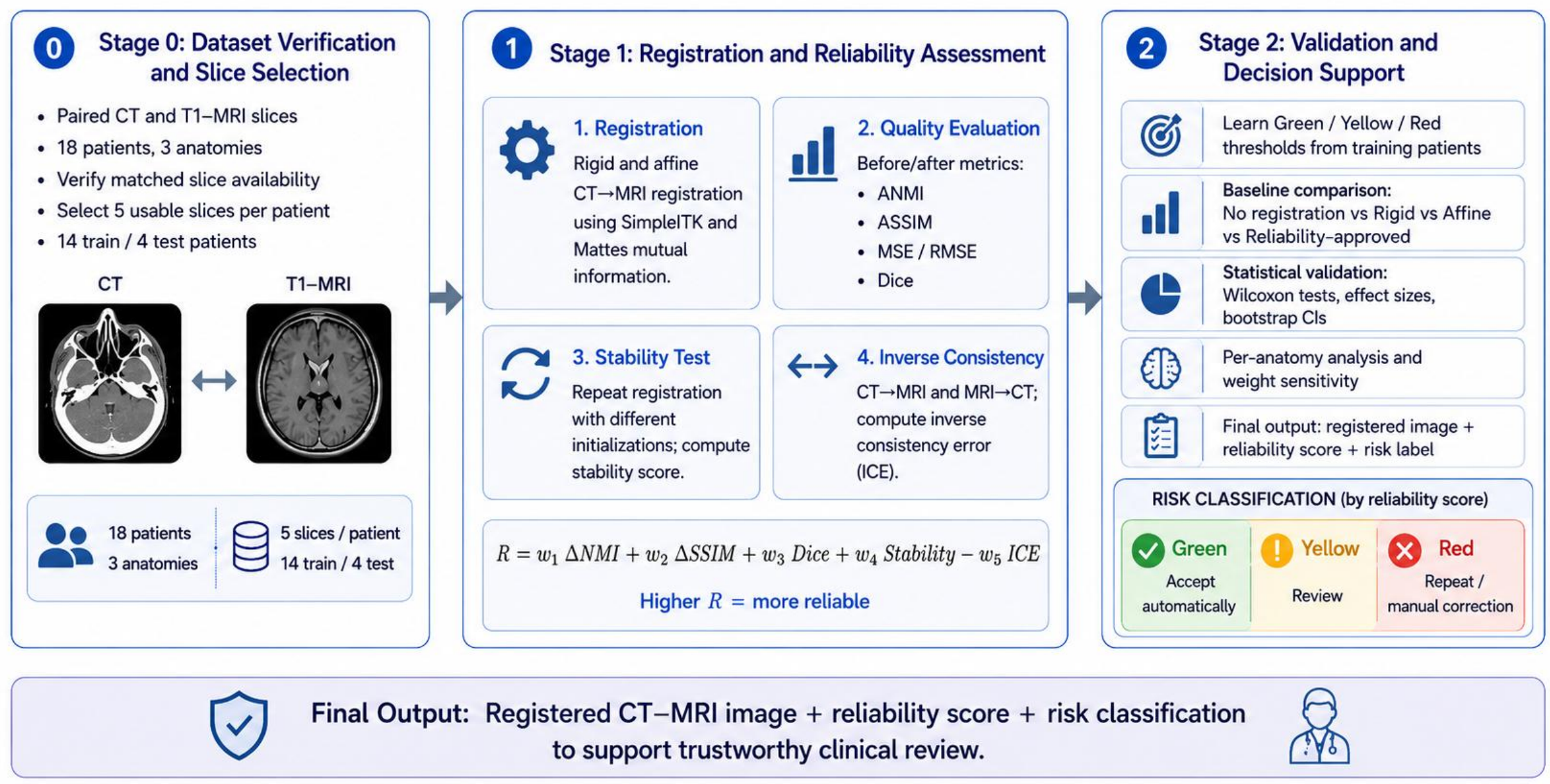

*Figure 1. Methodology overview of the reliability-aware quality-control framework for CT–MRI image registration. Stage 0 verifies dataset completeness and selects usable CT–T1-MRI slices across 18 patients (14 train / 4 test). Stage 1 performs rigid and affine registration, before/after quality evaluation (NMI, SSIM, MSE/RMSE, Dice), a stability test across repeated random initialisations, and an inverse consistency check (ICE), combining these into the reliability score $R = w_1\Delta NMI + w_2\Delta SSIM + w_3 Dice + w_4 Stability - w_5 ICE$. Stage 2 learns Green/Yellow/Red risk-classification thresholds from training patients, performs baseline comparison, statistical validation, and per-anatomy and weight-sensitivity analysis, and produces the final output: a registered image, reliability score, and risk label to support trustworthy clinical review.*

## 3.1 Dataset Verification and Slice Selection

This study uses the Paired CT and MRI Dataset for Medical Applications (Akon et al., Kaggle), comprising 18 patient volumes with co-registered CT, T1-weighted MRI, and T2-weighted MRI slices at matched anatomical positions. Patients span three anatomical regions: brain (9 patients), abdominal (8 patients), and neck (1 patient). For each patient, the CT, T1-MRI, and T2-MRI slice inventories are cross-checked, and the intersection of available CT and T1-MRI slice indices is computed to ensure only genuinely paired slices are used. Five slices are then selected uniformly across each patient's paired slice range, skipping the first and last edge slices to avoid near-empty body masks, yielding 90 total CT–MRI slice pairs (18 patients × 5 slices). A 14/4 training/test split is applied here (14 training patients: P01–P07, P10–P15, P18; 4 held-out test patients: P08, P09, P16, P17).

Verification confirmed that all 18 patients have fully paired CT and T1-MRI slices (389 paired slices total across the dataset), with no missing modalities or insufficient slice counts, and all patients were retained for the registration pipeline.

Full patient-level slice availability and selected slice indices for all 18 patients are provided in Table 1 (Section 4.1).

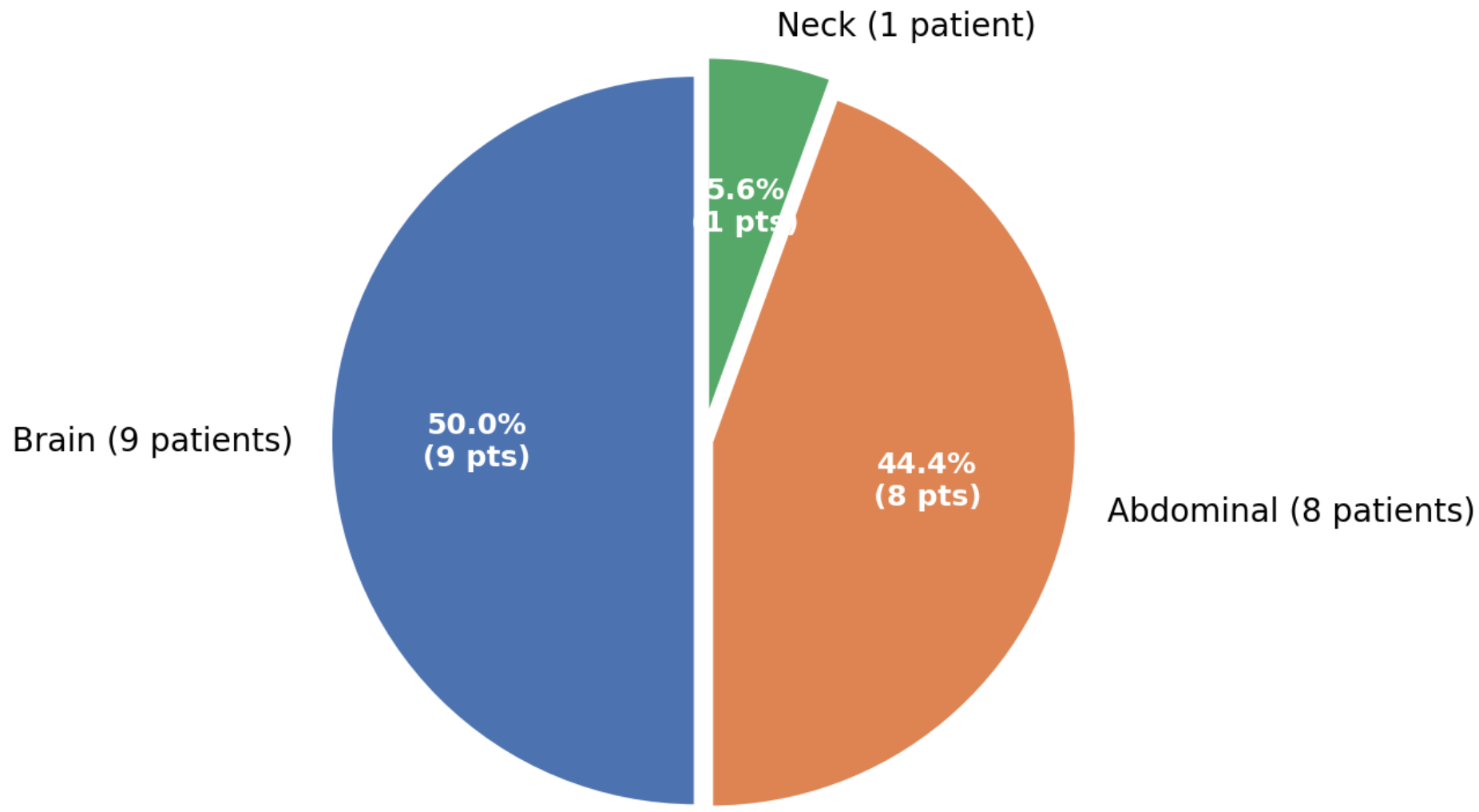

*Figure 2. Distribution of the 18 patients in the Paired CT and MRI Dataset for Medical Applications (Akon et al., Kaggle) across the three anatomical regions used in this study: brain (9 patients, 50.0%), abdominal (8 patients, 44.4%), and neck (1 patient, 5.6%). The single neck patient limits the statistical power of anatomy-specific conclusions for that region (see Section 5.5).*

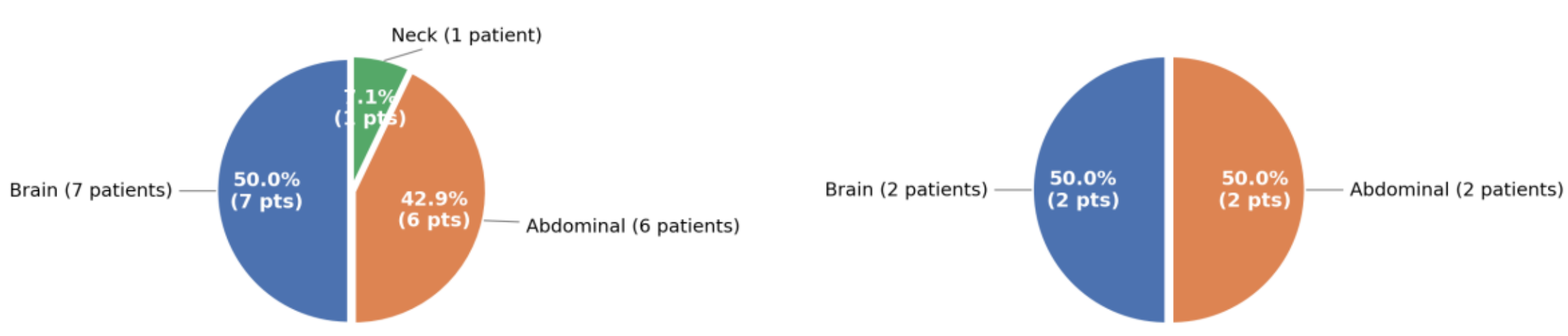

*Figure 3. Anatomical region distribution of the 14 training patients (brain: 7, 50.0%; abdominal: 6, 42.9%; neck: 1, 7.1%) and the 4 held-out test patients (brain: 2, 50.0%; abdominal: 2, 50.0%; neck: 0). The single neck patient (P18) falls in the training set only.*

## 3.2 Registration Procedure

For each selected CT–MRI slice pair, the CT image (moving) is registered to the T1-MRI image (fixed) using SimpleITK's intensity-based registration framework. Two transformation classes are evaluated independently: rigid (2-D Euler transform: rotation and translation) and affine (rotation, translation, scaling, and shearing). The similarity metric is Mattes mutual information, computed via random sampling of 50% of image pixels with 50 histogram bins, optimised using regular-step gradient descent (initial learning rate 1.0, minimum step $10^{-6}$, 200 iterations, relaxation factor 0.5) with optimiser scales set from the physical shift heuristic. Both CT and MRI images are normalised to [0, 1] prior to registration, and a binary body mask is derived by thresholding at intensity 0.1.

### 3.3 Quality Evaluation

Registration quality is assessed by computing four metrics on the CT–MRI pair both before and after registration: normalised mutual information (NMI), computed from the joint intensity histogram of the two images; structural similarity (SSIM), computed via the standard windowed implementation; mean squared error (MSE) and its root (RMSE); and Dice overlap between binary body masks derived from each image. The quality improvement attributable to registration is summarised as $\Delta NMI = NMI_{after} - NMI_{before}$ and $\Delta SSIM = SSIM_{after} - SSIM_{before}$.

### 3.4 Stability Testing

To assess whether a registration result is reproducible, the same CT–MRI pair is registered ten times from different random initialisations of the optimiser's metric sampling seed, while holding the transformation class and similarity metric fixed. The resulting transformation parameter vectors are collected, and a stability score is computed as one minus the mean normalised standard deviation of the parameters across initialisations (normalised by each parameter's observed range across the ten runs). A stability score near 1 indicates that registration converges to essentially the same transformation regardless of initialisation; a low stability score indicates sensitivity to starting conditions, a signal of an ill-conditioned or multi-modal optimisation landscape for that particular slice pair.

### 3.5 Inverse Consistency Analysis

Forward registration (CT→MRI) and backward registration (MRI→CT) are performed independently for each slice pair, following the inverse-consistency principle commonly used as an internal validity check in image registration [12,13]. The forward-registered CT is then transformed by the backward transform (round trip), and the inverse consistency error (ICE) is computed as the mean absolute pixel-intensity discrepancy between the original moving image and its round-trip reconstruction. ICE is used here as an internal reliability diagnostic that flags directional disagreement between forward and backward registration, rather than as a standalone, ground-truth-based measure of registration accuracy: a low ICE indicates that the forward and backward registrations are mutually consistent, while a high ICE signals that the two registration directions disagree, suggesting an unreliable alignment in at least one direction that warrants closer review.

### 3.6 Baseline Comparison

Registration outcomes are compared across four conditions on the same 90 slice pairs: no registration (raw CT–MRI alignment as acquired), rigid registration, affine registration, and reliability-approved registration (the subset of registrations classified as Green by the risk classifier). This comparison evaluates whether the reliability filter improves average quality relative to using rigid or affine registration unconditionally.

### 3.7 Statistical Validation

Paired Wilcoxon signed-rank tests assess whether rigid and affine registration produce statistically significant improvements in NMI and SSIM relative to no registration, computed separately for training patients, test patients, and the combined sample. The non-parametric Wilcoxon signed-rank test was chosen in preference to the paired t-test because it does not assume that the paired metric differences are normally distributed—an assumption difficult to justify at the present sample size—and is robust to outliers and the skewed distributions typical of image-similarity metrics. A separate paired test compares affine against rigid registration directly. Effect sizes are reported using Cohen's d for paired differences, and 95% confidence intervals on mean improvements are obtained via non-parametric bootstrap (10,000 resamples). Holm's sequential correction is applied across the family of statistical tests to control the family-wise error rate.

### 3.8 Per-Anatomy Analysis and Weight Sensitivity

Reliability outcomes are stratified by anatomical region (brain, abdominal, neck) to assess whether registration difficulty and reliability differ systematically across anatomies. Separately, a one-at-a-time weight sensitivity analysis varies each reliability-score weight (w1–w5) from 50% to 150% of its default value while holding the remaining weights fixed, recomputing R and the resulting Green/Yellow/Red distribution at each setting. This identifies which reliability components the score is most sensitive to, providing guidance for weight calibration in future deployments.

### 3.9 Evaluation Metrics and Reliability Formulation

For reproducibility, the evaluation metrics used in the before/after registration analysis are formalised below. Let X denote the fixed T1-MRI image, Y denote the CT image before or after registration, A and B denote the corresponding binary body masks, and N denote the number of pixels in a slice.

**Normalised Mutual Information (NMI).**

NMI measures statistical dependence between CT and MRI intensities and is suitable for multimodal registration because it does not require direct intensity correspondence between modalities.

$$H(X) = -\sum_{i} p_X(i) \log p_X(i)$$

$$\mathrm{NMI}(X,Y) = \frac{H(X) + H(Y)}{H(X,Y)}$$

$$\Delta\mathrm{NMI} = \mathrm{NMI}(X, Y_{\mathrm{after}}) - \mathrm{NMI}(X, Y_{\mathrm{before}})$$

**Structural Similarity Index (SSIM).**

SSIM measures structural agreement between the fixed MRI image and the transformed CT image using local luminance, contrast, and covariance terms.

$$\mathrm{SSIM}(X,Y)=\frac{(2\mu_X\mu_Y+C_1)(2\sigma_{XY}+C_2)}{(\mu_X^2+\mu_Y^2+C_1)(\sigma_X^2+\sigma_Y^2+C_2)}$$

$$\Delta\mathrm{SSIM}=\mathrm{SSIM}(X,Y_{\mathrm{after}})-\mathrm{SSIM}(X,Y_{\mathrm{before}})$$

**Mean Squared Error and Root Mean Squared Error.**

MSE and RMSE quantify pixel-wise residual discrepancy after normalisation to [0, 1], with lower values indicating smaller intensity disagreement.

$$\mathrm{MSE}(X,Y)=\frac{1}{N}\sum_{i=1}^{N}(X_i-Y_i)^2$$

$$\mathrm{RMSE}(X,Y)=\sqrt{\mathrm{MSE}(X,Y)}$$

**Dice Overlap.**

Dice overlap measures spatial agreement between the binary body masks extracted from the CT and MRI slices.

$$\mathrm{Dice}(A,B)=\frac{2|A\cap B|}{|A|+|B|}$$

**Stability Score.**

Registration stability is computed from repeated registrations under different random initialisations. If θ_j denotes the jth transformation parameter, the score penalises variability across repeated runs.

$$\mathrm{Stability}=1-\frac{1}{p}\sum_{j=1}^{p}\frac{\mathrm{sd}(\theta_j)}{\mathrm{range}(\theta_j)+\varepsilon}$$

**Inverse Consistency Error (ICE).**

ICE measures the discrepancy between the original CT image and its round-trip reconstruction after forward CT→MRI and backward MRI→CT registration.

$$\mathrm{ICE}=\frac{1}{N}\sum_{i=1}^{N}\left|Y_i-Y_i^{\text{round-trip}}\right|$$

**Reliability Score.**

After normalising all components to [0, 1], the final reliability score combines quality improvement, overlap, stability, and inverse consistency into a single quality-control criterion.

$$R=w_1\Delta\mathrm{NMI}+w_2\Delta\mathrm{SSIM}+w_3\mathrm{Dice}+w_4\mathrm{Stability}-w_5\mathrm{ICE}$$

Higher values of NMI, SSIM, Dice, Stability, and R indicate better alignment or reliability, whereas lower values of MSE, RMSE, and ICE are preferred.

The default weights are w = (0.25, 0.25, 0.20, 0.15, 0.15) for (ΔNMI, ΔSSIM, Dice, Stability, ICE) respectively, chosen to balance quality improvement, anatomical overlap, and process-reliability diagnostics. These weights reflect domain judgement rather than empirical optimisation; their influence on the reliability score is examined quantitatively in the sensitivity analysis (Sections 3.8 and 4.8). Each component is normalised to [0, 1] prior to combination using the 95th percentile of its empirical distribution across all 180 registrations as the upper reference point, ensuring the score reflects the actual range of values observed in this dataset rather than an arbitrary fixed scale. The resulting R is clipped to [0, 1], with higher values indicating greater reliability.

$$\mathbf{w} = (0.25, 0.25, 0.20, 0.15, 0.15)$$

R is mapped to a three-tier risk classification using thresholds learned exclusively from training-patient registrations: Green (R ≥ 65th percentile of training R, accept automatically), Yellow (R between the 30th and 65th percentile, flag for expert review), and Red (R below the 30th percentile, repeat registration or apply manual correction). These thresholds are fixed after training and applied unchanged to the four held-out test patients, mirroring the strict train/test separation used throughout this research programme.

$$\text{Risk}(R) = \begin{cases} \text{Green}, & R \geq Q_{0.65}^{\text{train}} \\ \text{Yellow}, & Q_{0.30}^{\text{train}} \leq R < Q_{0.65}^{\text{train}} \\ \text{Red}, & R < Q_{0.30}^{\text{train}} \end{cases}$$

**Algorithm 1. Reliability-Aware CT–MRI Registration Pipeline**

**Input:** Paired CT and T1–MRI slices
**Output:** Registered image, reliability score R, and risk label

1. Verify CT–MRI slice availability for each patient.
2. Select five paired slices per patient and split patients into training and held-out test sets.
3. For each selected CT–MRI slice pair:
   a. Normalise CT and MRI images.
   b. Register CT to T1-MRI using rigid and affine transformations.
   c. Compute before/after quality metrics: NMI, SSIM, MSE/RMSE, and Dice.
   d. Repeat registration under multiple initialisations and compute the stability score.
   e. Perform forward and backward registration and compute inverse consistency error.
   f. Combine ΔNMI, ΔSSIM, Dice, Stability, and ICE into reliability score R.
4. Learn Green/Yellow/Red thresholds using training patients only.
5. Apply fixed thresholds to training and held-out test registrations.
6. Compare no registration, rigid, affine, and reliability-approved results.
7. Perform statistical validation, per-anatomy analysis, and weight sensitivity analysis.

Algorithm 1 summarises the complete reliability-aware registration pipeline described in Sections 3.1–3.9, from dataset verification through registration, quality evaluation, stability and inverse-consistency diagnostics, reliability scoring, and threshold-based risk classification.

## 4. Results

This section reports results for each stage of the pipeline, mirroring the methodology structure.

### 4.1 Dataset Verification and Slice Selection

All 18 patients were confirmed to have complete, paired CT and T1-MRI slice inventories, with CT, T1-MRI, and T2-MRI slice counts matching exactly within each patient (ranging from 9 slices for the smallest volume to 49 for the largest). A total of 389 paired CT–T1-MRI slices were available across the dataset, from which 90 slices were selected (18 patients × 5 uniformly spaced slices), confirming the dataset was complete and ready for registration without any missing-modality issues.

***Table 1. Slice availability and selection summary for all 18 patients.***

| Patient | Anatomy | Split | CT Slices | T1-MRI Slices | T2-MRI Slices | CT∩T1 Paired | All Paired | Selected Slices | N Sel. | Status |
|---|---|---|---|---|---|---|---|---|---|---|
| Patient_01 | Brain | Train | 19 | 19 | 19 | 19 | 19 | [2, 6, 10, 14, 18] | 5 | OK |
| Patient_02 | Brain | Train | 20 | 20 | 20 | 20 | 20 | [2, 6, 10, 14, 19] | 5 | OK |
| Patient_03 | Brain | Train | 12 | 12 | 12 | 12 | 12 | [2, 4, 6, 8, 11] | 5 | OK |
| Patient_04 | Brain | Train | 9 | 9 | 9 | 9 | 9 | [2, 3, 5, 6, 8] | 5 | OK |
| Patient_05 | Brain | Train | 17 | 17 | 17 | 17 | 17 | [2, 5, 9, 12, 16] | 5 | OK |
| Patient_06 | Brain | Train | 20 | 20 | 20 | 20 | 20 | [2, 6, 10, 14, 19] | 5 | OK |
| Patient_07 | Brain | Train | 15 | 15 | 15 | 15 | 15 | [2, 5, 8, 11, 14] | 5 | OK |
| Patient_08 | Brain | Test | 20 | 20 | 20 | 20 | 20 | [2, 6, 10, 14, 19] | 5 | OK |
| Patient_09 | Brain | Test | 20 | 20 | 20 | 20 | 20 | [2, 6, 10, 14, 19] | 5 | OK |
| Patient_10 | Abd | Train | 19 | 19 | 19 | 19 | 19 | [2, 6, 10, 14, 18] | 5 | OK |
| Patient_11 | Abd | Train | 26 | 26 | 26 | 26 | 26 | [2, 7, 13, 19, 25] | 5 | OK |
| Patient_12 | Abd | Train | 49 | 49 | 49 | 49 | 49 | [2, 13, 25, 36, 48] | 5 | OK |
| Patient_13 | Abd | Train | 18 | 18 | 18 | 18 | 18 | [2, 5, 9, 13, 17] | 5 | OK |
| Patient_14 | Abd | Train | 42 | 42 | 42 | 42 | 42 | [2, 11, 21, 31, 41] | 5 | OK |
| Patient_15 | Abd | Train | 23 | 23 | 23 | 23 | 23 | [2, 7, 12, 17, 22] | 5 | OK |
| Patient_16 | Abd | Test | 20 | 20 | 20 | 20 | 20 | [2, 6, 10, 14, 19] | 5 | OK |
| Patient_17 | Abd | Test | 26 | 26 | 26 | 26 | 26 | [2, 7, 13, 19, 25] | 5 | OK |
| Patient_18 | Neck | Train | 14 | 14 | 14 | 14 | 14 | [2, 4, 7, 10, 13] | 5 | OK |

### 4.2 Registration Quality Evaluation

Across all 180 registrations (90 slice pairs × 2 methods), both rigid and affine registration produced measurable, consistent improvements in alignment quality relative to the unregistered baseline. Affine registration achieved a mean ΔNMI of 0.0118 and mean ΔSSIM of 0.0075, compared to 0.0078 and 0.0052 respectively for rigid registration, indicating that the additional degrees of freedom in the affine transform capture systematically more alignment improvement than rigid transformation alone. Mean Dice overlap after registration was high for both methods (0.851 for rigid, 0.855 for affine), reflecting that body masks were already substantially overlapping at acquisition. The absolute Dice gain attributable to registration was therefore modest, and the framework's primary contribution in this dataset lies in reliability classification rather than in producing large absolute Dice improvements.

*Representative Registration Examples.* To complement the aggregate statistics above, Table 2 and Figures 4–6 show qualitative before/after registration overlays for three representative test-patient slices spanning both available anatomies, alongside their corresponding quantitative quality changes. These representative examples are provided for visual illustration only; all quantitative evaluation, reliability scoring, and statistical validation were performed using the full 90-slice dataset.

***Table 2. Representative slice-level registration metrics for three held-out test-patient examples.*** ΔMSE = post-registration minus pre-registration MSE; negative values indicate a reduction in MSE (improvement), positive values an increase. Note that in multimodal CT–MRI registration, MSE can move in the opposite direction from NMI and SSIM due to the inherent intensity-scale mismatch between modalities; NMI and SSIM are therefore the primary quality indicators for this framework.

| Patient | Anatomy | Slice | Method | ΔNMI | ΔSSIM | ΔMSE |
|---|---|---|---|---|---|---|
| P08 | Brain | 10 | Rigid | +0.0037 | +0.0019 | +0.000275 |
| P08 | Brain | 10 | Affine | +0.0037 | +0.0032 | −0.000436 |
| P16 | Abdominal | 10 | Rigid | +0.0079 | +0.0068 | +0.000373 |
| P16 | Abdominal | 10 | Affine | +0.0128 | +0.0023 | +0.001750 |
| P17 | Abdominal | 13 | Rigid | +0.0014 | +0.0062 | −0.000545 |
| P17 | Abdominal | 13 | Affine | +0.0043 | +0.0151 | −0.001688 |

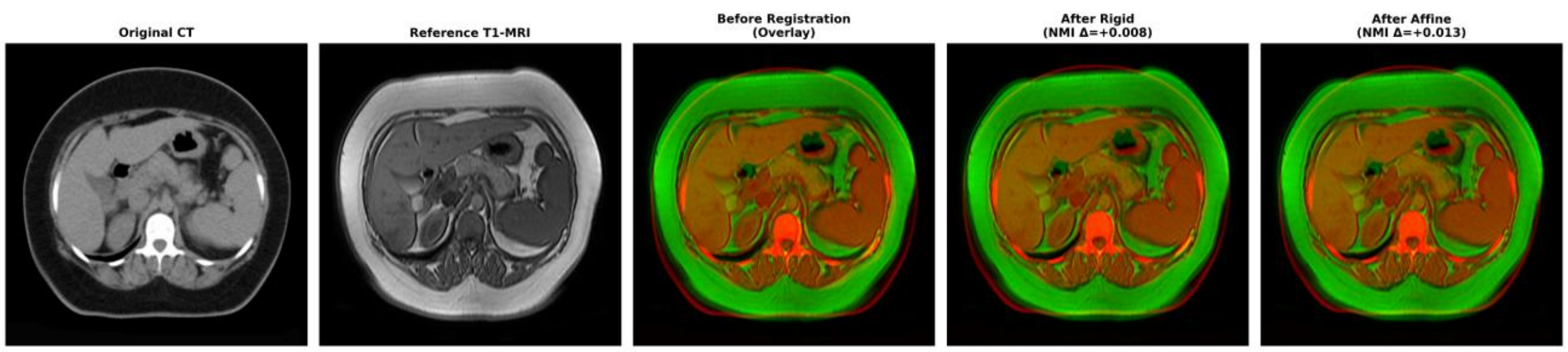

*Figure 4. Representative CT–MRI registration for patient P16 (abdominal, test set), slice 10. From left to right: original CT, reference T1-MRI, before-registration overlay, after-rigid overlay (NMI Δ = +0.008), and after-affine overlay (NMI Δ = +0.013). Red = CT, green = MRI, yellow = overlap; greater yellow overlap generally indicates improved spatial agreement.*

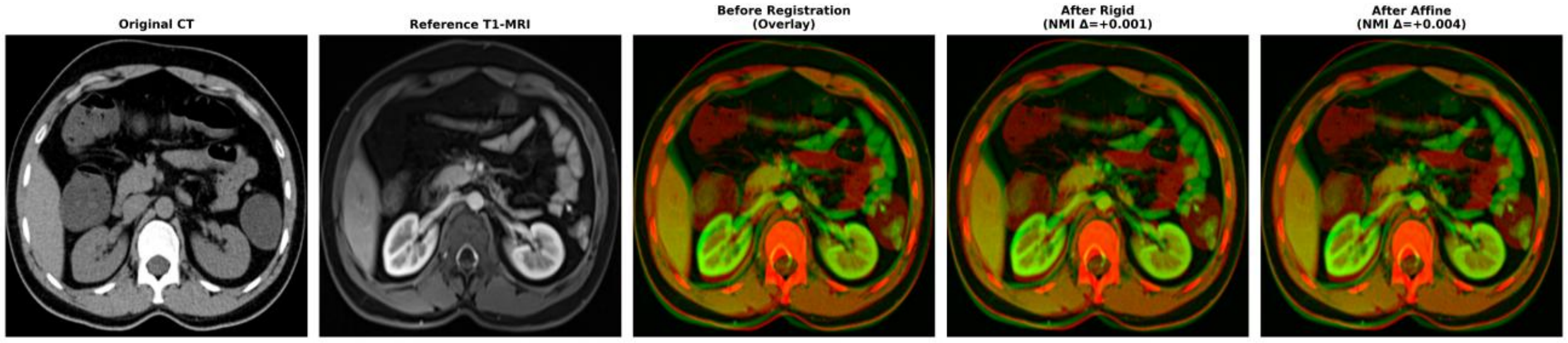

*Figure 5. Representative CT–MRI registration for patient P17 (abdominal, test set), slice 13. From left to right: original CT, reference T1-MRI, before-registration overlay, after-rigid overlay (NMI Δ = +0.001), and after-affine overlay (NMI Δ = +0.004). Red = CT, green = MRI, yellow = overlap. This case shows a smaller affine gain than P16, consistent with the variability in per-slice reliability captured by the framework.*

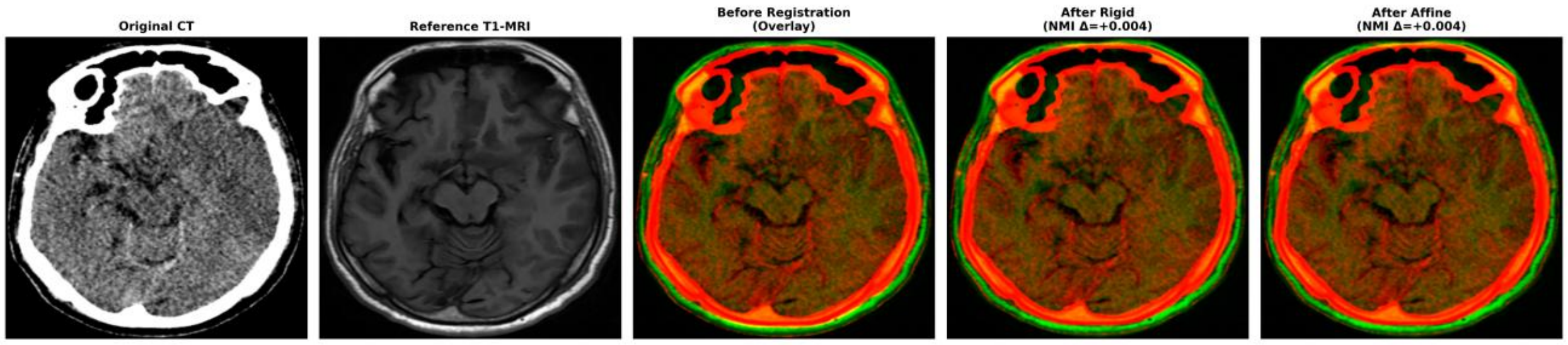

*Figure 6. Representative CT–MRI registration for patient P08 (brain, test set), slice 10. From left to right: original CT, reference T1-MRI, before-registration overlay, after-rigid overlay (NMI Δ = +0.004), and after-affine overlay (NMI Δ = +0.004). Red = CT, green = MRI, yellow = overlap. The smaller visible shift relative to the abdominal examples in Figures 4–5 reflects the brain's already close alignment at acquisition, consistent with the comparatively modest brain-anatomy gains reported in Section 4.7.*

## 4.3 Stability and Inverse Consistency Analysis

Mean stability scores were comparable between methods (0.693 for rigid, 0.696 for affine), indicating that both transformation classes converge consistently across repeated random initialisations for this dataset. Mean inverse consistency error was lower for rigid registration (0.026) than affine (0.038), consistent with the intuition that the additional affine degrees of freedom introduce more opportunity for forward–backward asymmetry, even though affine achieves better raw alignment quality.

## 4.4 Reliability Score and Risk Classification

Using data-driven normalisation (95th-percentile reference scaling) and thresholds learned from training-patient registrations (Green ≥ 0.375, Yellow ≥ 0.299), the reliability score distinguished a meaningful spread of registration trustworthiness across the dataset rather than classifying all cases identically. Affine registration produced a higher mean reliability score (R = 0.382) than rigid registration (R = 0.351) and a more favourable risk distribution.

*Table 3. Risk classification by registration method (90 slices per method).*

| Method | Green | Yellow | Red | Mean R |
|---|---|---|---|---|
| Rigid | 30 (33%) | 33 (37%) | 27 (30%) | 0.351 |
| Affine | 40 (44%) | 28 (31%) | 22 (24%) | 0.382 |

Affine registration yielded both a higher proportion of automatically acceptable (Green) registrations and a lower proportion requiring rejection (Red), suggesting that, despite its slightly worse inverse consistency, the additional flexibility of affine registration more often produces results that meet the combined reliability criterion.

## 4.5 Baseline Comparison

Table 4 compares mean NMI, SSIM, and MSE across four conditions: no registration, rigid registration, affine registration, and reliability-approved (Green-classified) registrations. Reliability-approved registrations achieved the highest NMI and SSIM of any condition, for both rigid and affine methods, confirming that the reliability filter successfully identifies the subset of registrations with the best alignment quality rather than simply correlating with arbitrary metric values. One metric-specific pattern warrants noting: R-Approved Rigid achieved a marginally higher mean NMI (1.1271) than R-Approved Affine (1.1226), a reversal of the general pattern in which affine outperforms rigid. This reflects the composition of the Green-classified subset for each method rather than an overall performance reversal; affine remains superior on SSIM and MSE, and on the full-dataset statistics reported in Section 4.6.

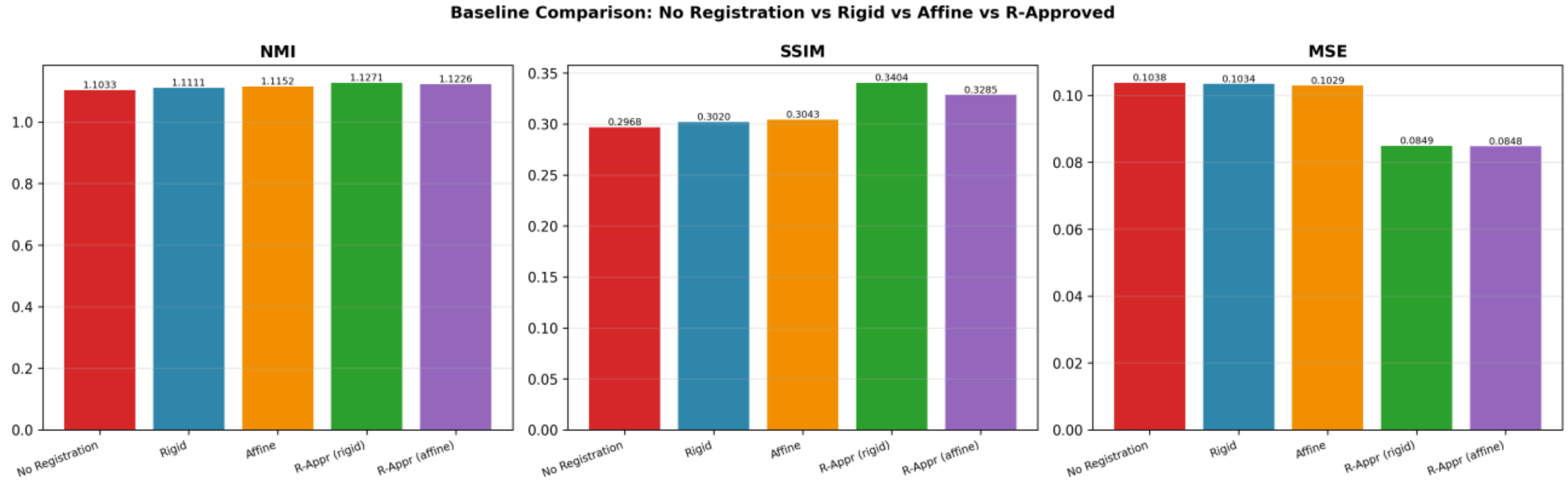

*Figure 7. Baseline comparison of mean NMI, SSIM, and MSE across no registration, rigid, affine, and reliability-approved (R-Appr) registration conditions, for both rigid and affine transformation classes. Reliability-approved registrations achieve the best score on every metric, with the largest gains visible in SSIM and MSE.*

*Table 4. Baseline comparison across registration conditions.*

| Condition | Mean NMI | Mean SSIM | Mean MSE |
|---|---|---|---|
| No Registration | 1.1033 | 0.2968 | 0.103766 |
| Rigid | 1.1111 | 0.3020 | 0.103405 |
| Affine | 1.1152 | 0.3043 | 0.102924 |
| R-Approved (Rigid) | 1.1271 | 0.3404 | 0.084932 |
| R-Approved (Affine) | 1.1226 | 0.3285 | 0.084798 |

## 4.6 Statistical Validation

Paired Wilcoxon signed-rank tests confirmed that both rigid and affine registration produce statistically significant improvements in NMI and SSIM relative to no registration, with large effect sizes for NMI (Cohen's d = 0.92 for rigid, d = 1.17 for affine, both $p < 10^{-16}$) and medium effect sizes for SSIM (d = 0.58 for rigid, d = 0.64 for affine, both $p < 10^{-8}$). These results held consistently across training patients, test patients, and the combined sample, indicating that the improvements generalise to held-out data rather than being an artefact of the training set. A direct paired comparison further confirmed that affine registration significantly outperforms rigid registration on NMI (d = 0.76, $p < 10^{-13}$) and, to a smaller degree, on SSIM (d = 0.36, $p < 0.01$). All reported comparisons remained significant after Holm's correction for multiple testing.

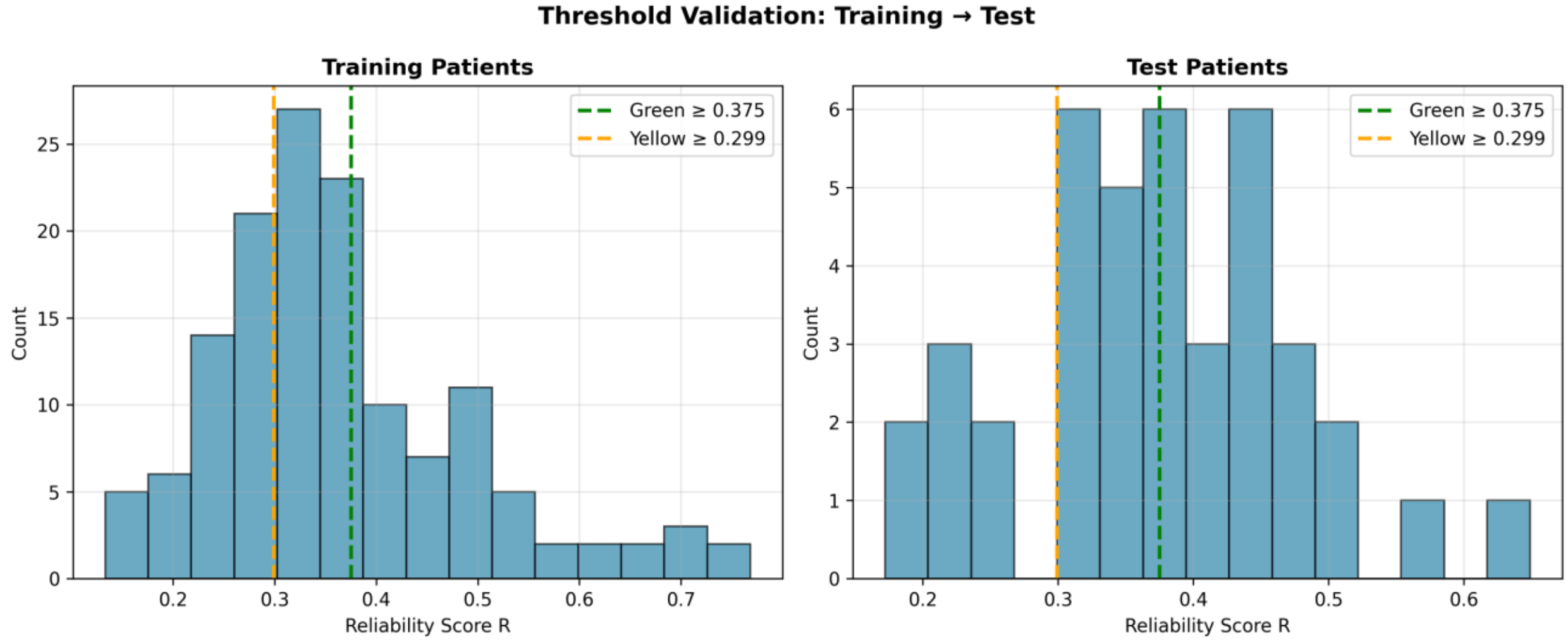

*Figure 8. Distribution of the reliability score R among training patients (left) and held-out test patients (right), with the Green ($R \geq 0.375$) and Yellow ($R \geq 0.299$) thresholds learned from training data overlaid on both panels. The test-patient distribution mirrors the shape and spread of the training distribution around both thresholds, supporting the validity of fixing the risk-classification cut-points on training data alone.*

***Table 5. Statistical validation summary (Wilcoxon signed-rank tests, all-patients sample).***

| Comparison | Mean Δ | Cohen's d | Effect | p-value | Significant |
|---|---|---|---|---|---|
| Rigid NMI vs No Reg | +0.0078 | 0.92 | Large | $<10^{-16}$ | Yes |
| Affine NMI vs No Reg | +0.0118 | 1.17 | Large | $<10^{-16}$ | Yes |

| Rigid SSIM vs No Reg | +0.0052 | 0.58 | Medium | $<10^{-8}$ | Yes |
|---|---|---|---|---|---|
| Affine SSIM vs No Reg | +0.0075 | 0.64 | Medium | $<10^{-8}$ | Yes |
| Affine vs Rigid (NMI) | +0.0041 | 0.76 | Medium | $<10^{-13}$ | Yes |
| Affine vs Rigid (SSIM) | +0.0022 | 0.36 | Small | <0.01 | Yes |

## 4.7 Per-Anatomy Analysis

Reliability outcomes differed substantially across anatomical regions. Abdominal registrations showed the strongest response to the registration and reliability framework, with affine registration achieving 73% Green classification and a mean reliability score of 0.464—the highest of any anatomy–method combination. Brain registrations, by contrast, showed comparatively modest improvement (22% Green under affine, mean R = 0.315), consistent with the brain images already being closely aligned at acquisition given the rigid, bony cranial geometry that constrains anatomical variation between CT and MRI scans of the same patient. Neck registrations showed intermediate reliability (20% Green for both methods, mean R between 0.326 and 0.357), with rigid slightly outperforming affine for this anatomy, plausibly reflecting the more constrained, less deformable neck anatomy where additional affine degrees of freedom provide limited benefit and may introduce instability. It should be noted, however, that the neck subset comprised only one patient and five slice pairs per method (n = 5); the statistical power of all neck-specific observations is therefore very limited, and these findings should be interpreted as exploratory only.

***Table 6. Per-anatomy reliability summary.***

| Anatomy | Method | Mean R | Green% | Yellow% | Red% |
|---|---|---|---|---|---|
| Brain | Rigid | 0.310 | 15.6 | 48.9 | 35.6 |
| Brain | Affine | 0.315 | 22.2 | 40.0 | 37.8 |
| Abdominal | Rigid | 0.397 | 55.0 | 22.5 | 22.5 |
| Abdominal | Affine | 0.464 | 72.5 | 20.0 | 7.5 |
| Neck | Rigid | 0.357 | 20.0 | 40.0 | 40.0 |
| Neck | Affine | 0.326 | 20.0 | 40.0 | 40.0 |

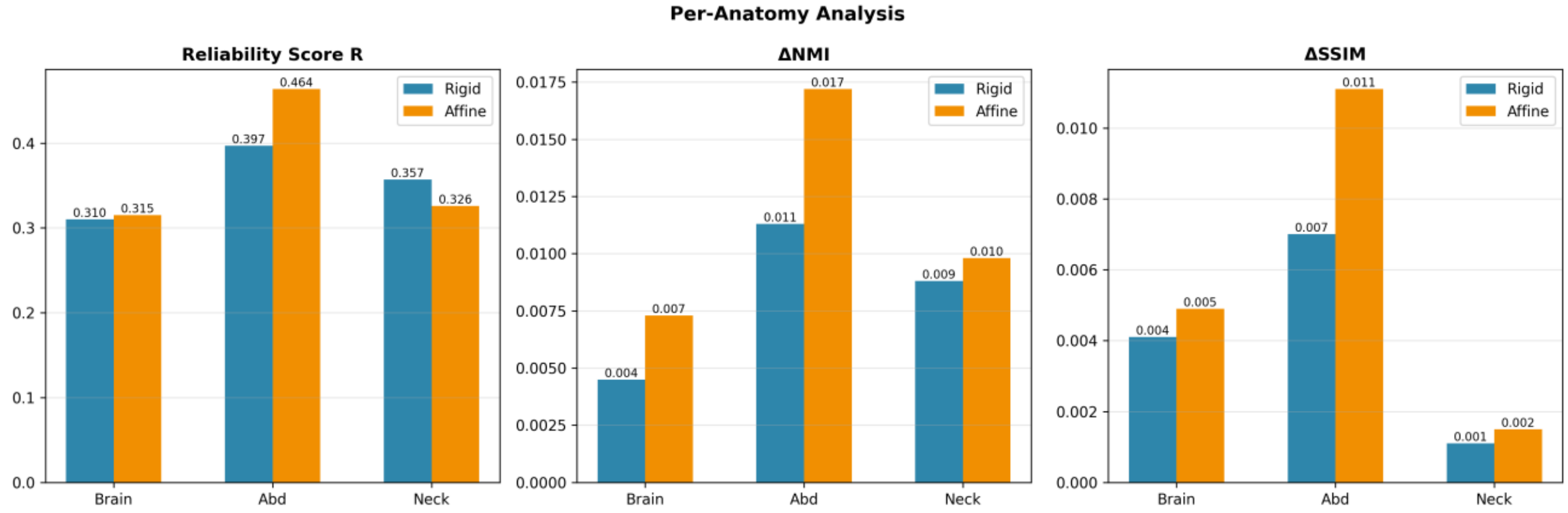

*Figure 9. Per-anatomy analysis showing mean reliability score R (left), mean ΔNMI (centre), and mean ΔSSIM (right) for rigid and affine registration across brain, abdominal, and neck anatomies. Abdominal registrations show*

*the largest gains and highest reliability under affine registration, while brain registrations show the smallest gains, consistent with their closer alignment at acquisition.*

## 4.8 Weight Sensitivity Analysis

Varying each reliability-score weight from 50% to 150% of its default value revealed that the score is most sensitive to the Dice overlap weight (w3): increasing it to 150% of default raised the Green classification rate from 39% to 76%, while reducing it to 50% lowered Green classification to 19%. The stability weight (w4) showed the second-largest sensitivity, with Green rates ranging from 26% to 60% across the tested range. The ΔNMI and ΔSSIM weights showed comparable, moderate sensitivity (Green ranging roughly 24–52%), while the inverse consistency weight (w5) showed an inverse relationship as expected—increasing its weight decreased the Green rate, from 48% at half-weight to 31% at one-and-a-half times the default weight. These results indicate that the overall risk classification is most strongly governed by anatomical overlap quality, suggesting that future deployments emphasising automated acceptance should pay particular attention to calibrating the Dice weight against domain-specific tolerance for residual misalignment.

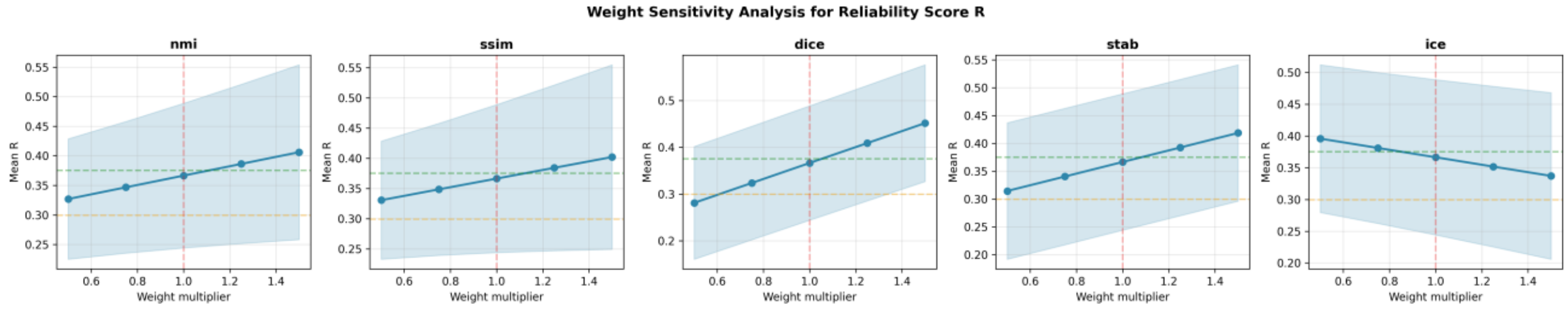

*Figure 10. Weight sensitivity analysis for the reliability score R, varying each component weight (NMI, SSIM, Dice, stability, ICE) from 50% to 150% of its default value while holding the others fixed (shaded bands show ±1 SD; dashed lines mark the Green/Yellow thresholds and the default weight multiplier). The Dice and stability weights show the steepest slopes, indicating the score is most sensitive to these two components.*

# 5. Discussion, Limitations, and Future Work

This section interprets the study's principal findings and delineates the boundaries of what the reliability-aware framework can and cannot claim. We first examine why affine registration outperforms rigid despite a higher inverse consistency error (Section 5.1), then consider how the value of registration varies by anatomy (Section 5.2) and the role of the Dice weight in risk classification (Section 5.3). We then set out the study's limitations (Section 5.4), outline directions for future work (Section 5.5), and compare our results with prior work (Section 5.6).

## 5.1 Why Affine Registration Outperforms Rigid Despite Higher Inverse Consistency Error

A notable finding is that affine registration achieves significantly better quality improvement and a more favourable risk distribution than rigid registration, despite exhibiting higher inverse consistency error. This apparent tension is resolved by recognising that the reliability score is a weighted combination of multiple, sometimes competing, diagnostics: affine registration's substantially larger gains in NMI, SSIM, and Dice overlap outweigh its modest disadvantage in inverse consistency under the default weighting. This illustrates a central motivation for combining multiple diagnostics into a single score rather than relying on inverse consistency or quality improvement alone—each diagnostic captures a different failure mode, and a registration method's overall trustworthiness depends on the balance between them. This balance is a deliberate property of the default weighting, which prioritises quality improvement and spatial overlap (combined weight 0.70) over the inverse-consistency term ($w_5 = 0.15$). Organisations whose pipelines depend heavily on forward–backward symmetry—for example, recursive or iterative registration schemes in which a transform is repeatedly composed with its inverse, or atlas-building workflows where consistency errors accumulate across many compositions—should increase $w_5$ accordingly. As a worked example, raising $w_5$ from 0.15 to 0.30 while rescaling the remaining weights proportionally roughly doubles the penalty that inverse consistency error imposes on the final score; this would narrow, and in borderline cases reverse, affine registration's advantage over rigid whenever the two methods differ mainly in consistency rather than in raw alignment quality.

### 5.2 The Anatomy-Dependent Value of Registration

The strong per-anatomy variation in reliability outcomes—particularly the high Green classification rate for abdominal registrations compared to the more modest rate for brain registrations—suggests that the marginal value of registration (and by extension, of careful reliability filtering) depends substantially on how well-aligned a given anatomy already is at acquisition. Brain imaging benefits from a rigid cranial vault that constrains relative anatomical position between sequential scans, leaving comparatively little residual misalignment for registration to correct, whereas abdominal anatomy is subject to greater positional and physiological variability between acquisitions, creating more opportunity for registration to provide meaningful quality improvement. This observation has practical implications: a quality-controlled imaging pipeline might reasonably calibrate anatomy-specific reliability thresholds, or anatomy-specific weight configurations, rather than applying a single global threshold uniformly.

### 5.3 The Role of the Dice Weight in Risk Classification

The weight sensitivity analysis identifying Dice overlap as the most influential reliability component carries a practical implication for deployment: because the Green classification rate is highly sensitive to this single weight, organisations adopting this framework should treat the Dice weight as the primary calibration lever when adjusting the framework's strictness, rather than attempting to tune all five weights simultaneously. This also suggests a natural extension for future work: learning the weights themselves from a validation set with expert-labelled ground-truth reliability assessments, rather than fixing them by domain judgement as in the present study. A practical caution is also warranted: placing disproportionate emphasis on the Dice overlap term may disadvantage patients with atypical body compositions, imaging artefacts affecting mask boundaries, or unusual scan positioning, where body-mask Dice may fail to reflect the quality of the underlying anatomical registration. Any deployment-stage calibration of the Dice weight should therefore be validated across a representative diversity of patient characteristics.

### 5.4 Limitations

Several limitations should be acknowledged:

1. The reliability score's components are normalised using percentile-based reference values derived from the same 180 registrations analysed here; while this provides a data-driven, reproducible calibration, the absolute reliability scores and thresholds are specific to this dataset and registration configuration and would require re-calibration for substantially different imaging protocols or patient populations.
2. The dataset is small (18 patients: 9 brain, 8 abdominal, 1 neck), which limits the precision of the percentile-based reference values and the generalisability of the learned thresholds; larger, multi-institutional datasets will be needed to establish population-level cut-points and to support a formal power analysis for reliable percentile estimation.
3. The Green/Yellow/Red thresholds, although learned from training data and validated on held-out test data, have not been validated against expert radiologist assessments of registration quality; the present study establishes statistical and quality-engineering validity but does not claim clinical validation of the risk categories themselves.
4. This study performs slice-level 2D registration rather than full volumetric 3D registration, chosen to keep the reliability formulation and the stability and inverse-consistency diagnostics tractable as a proof of concept; extending to 3D would enlarge the transformation parameter space and computational cost and may require threshold re-calibration.
5. Only rigid and affine registration are evaluated; deformable registration, which may be more appropriate for highly non-rigid anatomies such as the abdomen, was outside the present scope and represents a natural extension.
6. The dataset comprises a single MRI sequence (T1-weighted) registered to CT; extending the framework to T2-weighted and other sequences would test the generality of the reliability score across modality combinations with different contrast characteristics.

### 5.5 Future Work

Future work includes: extending the reliability framework to deformable registration, where stability and inverse consistency diagnostics become substantially more complex to compute but arguably more important given the larger parameter space; incorporating expert-labelled ground-truth reliability assessments to validate and potentially recalibrate the risk thresholds against clinical judgement; exploring learned, rather than fixed, component weights; and extending the per-anatomy analysis to a larger, more balanced sample of neck and other anatomies currently underrepresented in this dataset (a single neck patient limits the statistical power of anatomy-specific conclusions for that region). A concrete near-term priority is extension to full 3D volumetric registration, where inter-slice continuity, a larger transformation parameter space, and higher computational cost will all affect the stability and inverse-consistency diagnostics and are likely to require re-learning the risk thresholds. For deformable registration specifically, non-linear models such as B-spline free-form deformations or thin-plate splines introduce many local degrees of freedom; the stability metric will need to be adapted from global transform-parameter perturbation to spatially localised displacement-field analysis, and this adaptation should be developed and validated before any threshold re-learning is attempted.

## 5.6 Comparison with Prior Works

Table 7 situates the proposed framework relative to representative prior studies across six dimensions: registration type, quality metrics used, reliability or uncertainty assessment, per-case decision support, threshold learning strategy, and dataset characteristics. Studies were selected to span classical intensity-based methods, deep-learning registration, inverse-consistency literature, and registration quality-assessment work.

**Table 7. Comparison of the proposed framework with representative prior works in CT–MRI registration, quality assessment, and reliability engineering.**

| Study | Year | Reg. Type | Quality Metrics | Reliability / Uncertainty | Per-Case Decision | Threshold Learning | Dataset |
|---|---|---|---|---|---|---|---|
| Hill et al. [1] | 2001 | Rigid + affine (intensity-based) | NMI, CC | None | None | None | Brain MRI |
| Reuter et al. [27] | 2010 | Rigid, inverse-consistent | NMI, Dice | IC as optimisation constraint | None | None | Brain MRI |
| Avants et al. [28] (ANTs SyN) | 2011 | Symmetric diffeomorphic (deformable) | Dice, Jacobian | Symmetric IC as regularisation | None | None | Brain, lung |
| Khallaghi et al. [29] | 2021 | Non-rigid (deformable) | Dice, TRE | Rapid single-metric QA flag | Binary pass/fail only | Fixed heuristic | Interventional CT–US |
| Maintz & Viergever [8] | 1998 | Survey (intensity-based) | MI / NMI, CC | None | None | None | Multi-modal (review) |
| Losnegård et al. [30] | 2020 | Rigid (automated QA) | NMI, local metrics | Post-hoc auto accuracy flag | Binary per-case label | Fixed threshold | CT–MR head (18 patients) |
| Greer et al. [31] | 2023 | Deep learning, multi-step | Dice, TRE, ICE | IC by network construction | None | N/A (structural) | Brain, lung, knee |
| Eppenhof & Pluim [32] | 2019 | CNN-based | TRE, Dice | MC dropout uncertainty | Confidence score only | Not reported | Pulmonary CT |
| **Proposed Work** | **2026** | **Rigid + affine (classical, 2D slice-level)** | **NMI, SSIM, MSE/RMSE, Dice** | **Composite R score: ΔNMI + ΔSSIM + Dice + Stability − ICE** | **Three-tier traffic-light (Green/Yellow/Red)** | **Learned from training patients only** | **Paired CT–T1 MRI, 18 patients, 3 anatomies** |

The proposed framework is unique in combining all four reliability diagnostics — quality improvement, spatial overlap, stability, and inverse consistency — into a single interpretable score with data-learned risk thresholds. Prior work either reports no reliability assessment, uses a single post-hoc metric flag, or embeds inverse consistency as a training constraint rather than a deployment-time diagnostic. The traffic-light risk stratification with held-out threshold validation is, to the authors' knowledge, the first such system applied to CT–MRI registration quality control.

Drawing these findings together, the following recommendations summarise the practical steps for adopting the framework in a clinical setting:

| Key Recommendations for Clinical Deployment |
| --- |
| 1. Adopt affine registration as the default for automated acceptance: it produced more automatically acceptable (44% vs. 33% Green) and fewer rejected registrations than rigid. |
| 2. Operate the traffic-light workflow directly: auto-accept Green, route Yellow to expert review, and repeat or manually correct Red before any downstream use. |
| 3. Re-calibrate the percentile reference values and Green/Yellow thresholds on local data before go-live; the cut-points reported here are specific to this dataset and registration configuration. |
| 4. Use the Dice-overlap weight as the primary lever for tuning strictness, and validate any change across a representative diversity of patient characteristics. |
| 5. Increase the inverse-consistency weight ($w_5$) for recursive, iterative, or atlas-building pipelines that depend on forward–backward symmetry. |
| 6. Set anatomy-specific expectations: automation yield is high for abdominal registration and lower for already-aligned brain; validate thresholds against expert radiologist consensus before clinical use. |

## 6. Conclusion

We have presented a reliability-aware framework for CT–MRI registration that combines quality improvement, anatomical overlap, registration stability, and inverse consistency into a single, interpretable reliability score, mapped to an actionable Green/Yellow/Red risk classification with thresholds learned from training data and validated on held-out test patients. Applied to an 18-patient paired CT–MRI dataset, the framework demonstrated that affine registration significantly outperforms rigid registration in both quality improvement and reliability classification, that reliability-approved registrations achieve the best alignment quality of any evaluated condition, and that registration reliability varies substantially and systematically across anatomical regions. The weight sensitivity analysis further identified anatomical overlap (Dice) as the most influential component of the reliability score, providing practical guidance for calibrating the framework in future deployments. By providing a per-case, statistically validated reliability signal rather than aggregate quality metrics alone, this framework offers a reproducible foundation for quality-controlled, reliability-aware multimodal registration pipelines, while we explicitly caution that the risk thresholds established here reflect statistical and quality-engineering validation rather than clinical validation against expert assessment, and should be recalibrated before deployment in any clinical decision-support context.

## Declarations

**Data Availability.** The dataset analysed in this study, the Paired CT and MRI Dataset for Medical Applications (Akon et al.), is publicly available on Kaggle. No new patient data were generated during this study.

**Code Availability.** The source code implementing the reliability-aware registration framework and the analyses reported in this paper is available from the author upon reasonable request.

**Ethics Approval.** This study used a publicly available, fully de-identified imaging dataset and did not involve any new experimentation on human participants or animals; institutional review board approval was therefore not required.

**Funding.** This research received no specific grant from any funding agency in the public, commercial, or not-for-profit sectors.
**Conflict of Interest.** The author declares no competing interests.
**Author Contributions.** N.A. is the sole author and was responsible for the study conception and design, methodology and software implementation, formal analysis, and writing of the manuscript.